\def\year{2021}\relax
\documentclass[letterpaper]{article} 
\usepackage{aaai21}  
\usepackage{times}  
\usepackage{helvet} 
\usepackage{courier}  
\usepackage[hyphens]{url}  
\usepackage{graphicx} 
\usepackage{natbib}  
\usepackage{caption} 
\usepackage{amsmath,amssymb, bbm}
\usepackage{booktabs}
\usepackage{multirow}
\usepackage{cleveref}
\usepackage{algorithm}
\usepackage[noend]{algpseudocode}
\usepackage{subcaption}
\usepackage{vector}
\usepackage{siunitx}

\usepackage[switch]{lineno}

\newcommand{\Sspace}{\mathcal{S}}
\newcommand{\Aspace}{\mathcal{A}}
\newcommand{\Ospace}{\mathcal{O}}

\renewcommand{\vec}[1]{\vect{#1}}
\newcommand{\mat}[1]{\vect{#1}}

\title{Improved POMDP Tree Search Planning with Prioritized Action Branching}
\author {
        John Mern,\textsuperscript{\rm 1}
        Anil Yildiz, \textsuperscript{\rm 1}
        Larry Bush \textsuperscript{\rm 2} 
        Tapan Mukerji \textsuperscript{\rm 3}
        Mykel J. Kochenderfer \textsuperscript{\rm 1}\\
}
\affiliations {
    \textsuperscript{\rm 1}Stanford University, Department of Aeronautics and Astronautics, 496 Lomita Mall, Stanford, CA 94305 \\
    \textsuperscript{\rm 2}General Motors, Research and Development, Warren, MI \\
    \textsuperscript{\rm 3}Stanford University, Department of Energy Resources Engineering, 367 Panama Street, Stanford, CA 94305 \\
    \{jmern91, yildiz, mukerji, mykel\}@stanford.edu \\
    bushL2@alum.mit.edu
}
\begin{document}

\maketitle
\begin{abstract}
Online solvers for partially observable Markov decision processes have difficulty scaling to problems with large action spaces. 
This paper proposes a method called PA-POMCPOW to sample a subset of the action space that provides varying mixtures of exploitation and exploration for inclusion in a search tree. 
The proposed method first evaluates the action space according to a score function that is a linear combination of expected reward and expected information gain. 
The actions with the highest score are then added to the search tree during tree expansion. 
Experiments show that PA-POMCPOW is able to outperform existing state-of-the-art solvers on problems with large discrete action spaces. 
\end{abstract}

\section{Introduction}

Sequential decision making problems under uncertainty are often modeled as partially observable Markov decision processes (POMDPs)~\cite{littman1995}. 
A solution to a POMDP is a policy that maps a belief over the state of the environment to an optimal action that maximizes the sum of discounted rewards over a series of steps. 
Solving POMDPs exactly is generally intractable and has been shown to be \textit{PSPACE-complete} for finite horizons~\cite{papadimitriou1987}.
Therefore, a variety of offline and online approximate solution methods have been proposed~\cite{kochenderfer2015}. 

Offline solvers compute the full policy before any action is taken, and are typically effective at small to moderately sized POMDPs~\cite{ross2008}. 
Monte-Carlo methods using point-based belief space interpolation were initially explored~\cite{thrun1999}.
Many advanced solvers now use point-based value iteration to learn an approximation to the belief value function from a finite-set of belief points~\cite{kurniawati2008}.
However, because offline solvers compute policies over the entire belief space, they are typically not viable for large problems.

Several online planners have been developed by adapting Monte-Carlo Tree Search (MCTS) for partially observable environments. 
Because online planners only reason about beliefs reachable from the current belief, they can typically be applied to much larger problems~\cite{Silver2010, somani2013,sunberg2018}.

The POMCP algorithm~\cite{Silver2010} adapted UCT search~\cite{kocsis2006} by using generative models and sampling states from an unweighted particle set to search over action-observation trajectories.
The DESPOT algorithm~\cite{somani2013} takes a similar approach, using a deterministic generative model to reduce the tree search variance. 
The ABT  algorithm~\cite{kurniawati2016} was proposed to improve planning speed by reusing part of the previous belief step search tree. 

\begin{figure}[t]
    \centering
    \includegraphics[width=1.0\columnwidth]{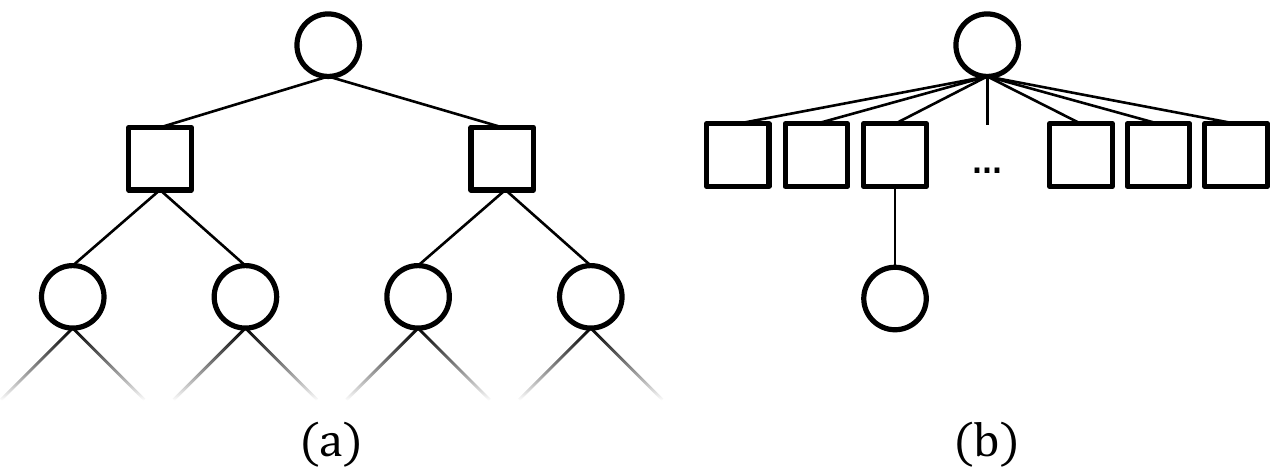}
    \caption{MCTS trees. (a) A deep search tree with a small action space. Action nodes (squares) are sampled frequently. 
    (b) A shallow search tree with a large action space.}
    \label{fig:tree_example}
\end{figure}

Existing online methods may still fail when the action space of the problem is very large, such as in large-scale route planning or robotic control.
During tree search, the probability of sampling a given action from a large space is very low, resulting in wide, shallow search trees~\cite{sunberg2018}, as shown in \Cref{fig:tree_example}. 
In MCTS methods, shallow trees provide poor estimates of the action values~\cite{Silver2010}.

To scale to problems with larger action and observation spaces, the POMCPOW and PFT-DPW algorithms~\cite{sunberg2018} introduce double progressive widening (DPW) to POMCP.
Progressive Widening~\cite{couetoux2011} was introduced to scale MCTS methods to large discrete and continuous spaces by dynamically limiting the number of action nodes added during tree expansion. 
\textit{Double} progressive widening applies the progressive widening to both the action and observation spaces.
DPW has been shown to be sensitive to the order nodes are selected for addition~\cite{browne2012} and has limited effect on scaling to large  or multidimensional action spaces.

We propose a method to select a subset of the most promising actions from the full action space.
Exploration in the tree is then limited to this smaller subset.
We select this subset according to a score function that evaluates each action's expected one-step reward and information gain. 
We provide formulations of this score for various reward functions and belief distributions. 
The method is implemented as an extension to POMCPOW. 
Experiments show that the proposed algorithm is able to outperform existing solvers on tasks with very large action spaces.

\section{Background} \label{sec:Background}

POMDPs represent sequential decision problems with state uncertainty~\cite{kochenderfer2015}. 
A POMDP is defined by the tuple $(\Sspace, \Aspace, \Ospace, Z, T, R, \gamma)$, where $\Sspace$, $\Aspace$, and $\Ospace$ are the state, action, and observation spaces, respectively. 
The transition model $T(s' \mid s,a)$ gives the probability of transitioning from state $s$ to state $s'$ after taking action $a$.
The reward function $R(s,a)$ specifies the immediate reward obtained after taking action $a$ at state $s$.
The observation model $Z(o \mid s,a,s')$ gives the probability of receiving observation $o$ in state $s'$ given that action $a$ had been taken in state $s$. The discount factor is $\gamma \in \left[0,1\right]$.

Because the state is unknown in a POMDP, it is common to maintain a probability distribution over the state, called the belief $b$.
The belief is updated each time the agent takes an action $a$ and receives an observation $o$, typically using a Bayesian update.

The action-value function $Q(b,a)$ is the expected sum of discounted future rewards when taking action $a$ at belief state $b$ and acting optimally for every following step.
Many POMDP solvers and planners operate by developing estimates of the action-value function over the action space and returning the action with the highest value.

Monte-Carlo Tree Search (MCTS) is often used for solving POMDPs. 
MCTS incrementally builds a tree of alternating layers of observation and action nodes by running many random simulations over the tree.
Simulations proceed from the root by selecting actions according to a given search policy and receiving observations from a generative model.
When a new observation is encountered, the observation node and its action node children are added to the tree.
This process is referred to as \textit{tree expansion}.

An estimate of $Q(b,a)$ is maintained for each action node.
Each time a node is visited, the value estimate is updated. 
More visits to a node generally improves the accuracy of the value function estimate~\cite{auer2002}.
POMDPs with large action or observation spaces typically result in wide, shallow trees. 
In these cases, the value function estimates may be poor.

\section{Proposed Method} \label{sec:Method}
The objective of this work is to improve the ability of online POMDP tree search planners to scale to very large action spaces. 
The challenge to tree search methods posed by large action spaces is the high branching factor they introduce in search tree expansion. 
To overcome this, we propose a method to select a small, informed subset of the action space for node expansion.
Our method uses the belief and reward function to select a subset of actions that balance exploration and reward. 

\subsection{Action Score Function}
In order to limit the branching factor for problems with large action spaces, only a subset of actions may necessarily be added to the search tree at each node. 
To select this subset, we propose ranking the actions according to a score function and including the actions with the highest scores in the subset.
We propose the \textit{action score} function that evaluates both the expected reward and expected information gain of an action
\begin{equation}
k(a, b; \lambda) = \mathbb{E}_{s \sim b}\left[r(s,a)\right] + \lambda I(b,a) \label{eq:score}
\end{equation}
where $a$ is the action, $b$ is the state belief distribution, $I(b,a)$ is an information gain term, and $\lambda$ is a weighting parameter. 
Including expected information gain is important to allow the planner to explore non-myopic trajectories.


Information gain cannot be efficiently computed for general distributions~\cite{frazier2018}.
Assuming the belief is unbiased at each time-step, we approximate the information gain as
\begin{align}
    IG(b', b) &= h(b) - h(b')~\label{infogain_1} \\
    &\approx \frac{1}{2}\log((2\pi e)^d|\Sigma_b|) - \frac{1}{2}\log((2\pi e)^d|\Sigma_{b'}|)~\label{infogain_2} \\
    &= \frac{1}{2}\big(\log(|\Sigma_b|) - \log(|\Sigma_{b'}|)\big)~\label{infogain_3} \\
    &= \frac{1}{2}\big(\mathrm{Tr}(\log(\Sigma_b)) - \mathrm{Tr}(\log(\Sigma_{b'}))\big)~\label{infogain_4}
\end{align}
where $\Sigma_b$ is the covariance of the belief distribution $b$ and $\Sigma_{b'}$ is the covariance of the updated belief $b'$ after having taken action $a$ and received an observation. 
\Cref{infogain_2} follows from the upper bound on differential entropy for continuous variables~\cite{cover2006}, and \cref{infogain_4} follows from the definition of the determinant of a matrix as $|X| = \exp[\mathrm{Tr}(\log(X))]$.

The proposed information gain score component is then 
\begin{equation}
    I(a,b) = \mathrm{Tr}(\log(\Sigma_b))-\mathrm{Tr}(\log(\mathbb{E}_{b'}\left[\Sigma_{b'}\right]))
\end{equation}
where the expectation is taken over the $b'$ term because the observation received upon taking action $a$ in belief $b$ is not necessarily known. 

Because the score will be evaluated for each action node, its evaluation needs to be fast.
The terms of the action score can be calculated exactly for some special cases, which are defined in the remainder of this section. 
In cases where analytical solutions are unavailable, approximations, for example by local linearization, may be used. 

\subsubsection{Expected Reward Term}
The expected reward term is generally defined as 
\begin{equation}
    \mathbb{E}_b\left[r(s,a)\right] = \int_s r(s,a) b(s) ds
\end{equation}
where $r(s,a)$ is the known reward function.


The expected reward can be exactly calculated for finite discrete state spaces by weighted summation over the belief.
For continuous cases, analytical solutions to the integral exist for special combinations of reward function and belief distribution. 
For instance, reward functions that are linear with respect to the state allow an analytical solution to be found for any distribution with a known first moment.
That is, given a reward function of the form
\begin{equation}
    r(s,a) = \vec{s}^T\vec{A}(a) + c(a)
\end{equation}
where $\vec{s}$ is a vector representation of the state, and the vector $\vec{A}$ and scalar $c$ are functions of the action $a$, the expected reward can be calculated as
\begin{equation}
    \mathbb{E}_b\left[r(s,a)\right] = \vec{\mu}_s^T\vec{A}(a) + c(a)
\end{equation}
where $\vec{\mu}_s$ is the mean of the state belief distribution. 
Given a Gaussian distribution belief, reward functions that are linear, quadratic, cubic, and quartic have known solutions~\cite{petersen2012}. 

\subsubsection{Expected Information Gain Term}
For the information gain term, we can define the expectation as 
\begin{align}
    \mathbb{E} & \left[\Sigma_{b'}\right] = \int_o \Sigma_{b'} P(o \mid a) do \\
    &= \int_o \Sigma_{b'} \int_s P(o \mid s, a) b(s) ds \ do \\
    &= \int_o \Sigma_{b'} \mathbb{E}_b[P(o \mid a)] do
\end{align}
where $b'$ is the updated belief after taking action $a$ and receiving observation $o$.
For the special case of finite discrete state and observation spaces, the information gain term can be calculated exactly as a summation over the distributions.
In the continuous domain, analytical solutions can be derived for special cases. 

For the common choice of a linear-Gaussian observation model and Gaussian belief, $\mathbb{E}_b[P(o \mid a)]$ is Gaussian.
The mean of the resulting distribution is $\mat{B}\vec{s} + \vec{d}$ and the covariance is $\mat{B}\Sigma_s \mat{B}^T + \Sigma_o$~\cite{kalman1960}, where $\Sigma_s$ is the belief covariance and $\Sigma_o$ is the observation noise covariance, which may be a function of the action.
The $\mat{B}$ matrix and $\vec{d}$ vector are the the affine transition matrix and bias vector, respectively. 

Given the Gaussian $\mathbb{E}_b[P(o \mid a)]$, the remainder of the solution of $ \mathbb{E} \left[\Sigma_{b'}\right]$ depends upon the function relating $\Sigma_o$ and the action $a$.
As with the reward, up to quartic relationships have known solutions.

Modeling the belief with a Gaussian Process (GP)~\cite{rasmussen2005} 
also gives an analytical solution.
Given a Gaussian Process, the belief distribution is a Gaussian distribution whose parameters are calculated from the posterior of the GP.
If we limit the observable points of the GP to elements of the state, the variance reduction is proportional to the marginal variance of the observed element.
Information gain then reduces to
\begin{equation}
    I(a,b) \propto \Sigma_b[o_x] - \sigma_o
\end{equation}
where $o_x$ is the index into the covariance matrix of the observed point, and $\sigma_o$ is the marginal observation noise. 

\subsection{Action Selection} \label{subsec: Action_Select}

The score function is used to select the best actions to add to the search tree at expansion steps.
We propose two methods for selecting actions using the score.
The subset method is proposed for planners, such as POMCP, that add all action nodes in a single step.
The prioritization method is proposed as an option for planners that use action progressive widening, such as POMCPOW.
For progressive widening solvers, either of the presented approaches may be used.

In the subset method, we propose only adding a subset of the total action space containing the highest scoring actions $\tilde{\mathcal{A}}$.
To select the subset, we can define a set of non-negative numbers $\Lambda$ such that $|\Lambda| \ll |\mathcal{A}|$.
We then define our subset to be $\tilde{A} \leftarrow \{a_i, ... , a_N\}$ where
\begin{equation}
  a_i \leftarrow \arg\max_{a \in \mathcal{A}} k(a,b;\lambda_i)
\label{eq:Pareto}
\end{equation}
and $\ \lambda_i \in \Lambda$.
Selecting actions in this way will result in a subset that exists along the Pareto frontier of the multi-objective action score, balancing reward gain and exploration.
%

In the prioritization method, a single value of $\lambda$ is set and the action with the highest score is added each time the tree is expanded.
No upper limit on the total number of added nodes is imposed with the aim of preserving the asymptotic convergence of behavior of the planner~\cite{kocsis2006, Silver2010}. 

These selection methods can be applied to any tree-search method which explicitly branches on the actions. 
The only additional information required is the score function, which only requires the explicit reward function to formulate.
The methods will only be effective on tasks with non-sparse rewards however, as the reward function is directly used to select actions for branching.
For tasks in which this is not the case, a shaped reward function may be used in place of the task reward in evaluating the action score.

The action score is only used in selecting actions to branch, not in chosing trajectories to explore or in node evaluation during rollout.
As a result, any optimality guarantees provided by the tree search algorithm, such as those provided by UCT~\cite{couetoux2011}, remain valid. 

\subsection{Algorithm}
We implement the proposed method in an extension of the POMCPOW online solver~\cite{sunberg2018} that can scale to very large problems.
We call this algorithm Prioritized Action POMCPOW (PA-POMCPOW). 
The same notation as in the original work is used.
Only the additions and modified functions are presented here due to space constraints.

The main modification is to the \textsc{ActionProgWiden} function presented in~\Cref{alg: ACTWIDE}, which defines the progressive widening procedure for the action space. 
The action selection procedure is defined in~\Cref{alg: SELACTS}.
Additionally, the PA-POMCPOW search tree is augmented with a set $E$, which stores the ordered action subset for each observation node. 
This set is maintained in order to minimize the number of calls required to \textsc{SelectActions}, as scoring the entire action space can be computationally expensive.

The new progressive widening step is defined such that only actions from the action subspace $\tilde{\mathcal{A}}$ are added to the tree. 
In addition to the history $h$, \textsc{ActionProgWiden} also takes the action-score function $k$, and $\Lambda$ as arguments.
The $\tilde{\mathcal{A}}$ set can be selected using either procedure described in~\Cref{subsec: Action_Select}. 
To use the subset method, a set of information gain weights is passed in for $\Lambda$. 
To use the prioritization method, a set with a single weight $\{\lambda\}$ is passed.
\begin{algorithm}[ht] 
\caption{\textsc{ActionProgWiden} Function}
\begin{algorithmic}[1]\label{alg: ACTWIDE}
\Function{\textsc{ActionProgWiden}}{$h, k, \lambda, \Lambda$}
\If{$h \notin E$}
    \State $\tilde{\mathcal{A}} \leftarrow \textsc{SelectActions}(h, \mathcal{A}, k, \Lambda)$
    \State $E(h) \leftarrow \tilde{\mathcal{A}}$
\EndIf
\If{$\|C(h)\| \leq k_aN(h)^{\alpha_a}$ and $E(h) \neq \emptyset$}
    \State $\tilde{\mathcal{A}} \leftarrow E(h)$
    \State $a \leftarrow \textsc{Next}(\tilde{\mathcal{A}})$ \Comment{Get next element from set} 
    \State $C(h) \leftarrow C(h) \cup {a}$
    \State $E(h) \leftarrow \tilde{\mathcal{A}}\setminus\{a\}$
\EndIf
\\
\Return $\arg\max_{a\in C(h)} Q(h, a) + c\sqrt{\frac{\text{log}N(h)}{N(ha)}}$
\EndFunction
\end{algorithmic}
\end{algorithm}
\begin{algorithm} 
\caption{\textsc{SelectActions} Function}
\begin{algorithmic}[1]\label{alg: SELACTS}
\Function{\textsc{SelectActions}}{$b, k, \Lambda$}
\State $\tilde{\mathcal{A}} \leftarrow \emptyset$
\State $\mathcal{A}' \leftarrow \mathcal{A}$
\If{$|\Lambda| = 1$}
    \State $\lambda \leftarrow \Lambda_0$
    \While{$|\mathcal{A}'|>0$}
        \State $a \leftarrow \arg\max_{a \in \mathcal{A}'} k(a,b;\lambda)$ 
        \State $\tilde{\mathcal{A}} \leftarrow \tilde{\mathcal{A}} \cup \{a\}$
        \State $\mathcal{A}' \leftarrow \mathcal{A}'\setminus \{a\}$
    \EndWhile
\Else
    \For{$\lambda \in \Lambda$}
        \State $a \leftarrow \arg\max_{a \in \mathcal{A}'} k(a,b;\lambda)$ 
        \State $\tilde{\mathcal{A}} \leftarrow \tilde{\mathcal{A}} \cup \{a\}$
        \State $\mathcal{A}' \leftarrow \mathcal{A}'\setminus \{a\}$
    \EndFor
\EndIf
\\
\Return $\tilde{\mathcal{A}}$
\EndFunction
\end{algorithmic}
\end{algorithm}

Different selections of $\Lambda$ lead to different solver behavior.
In general, larger sets require a higher number of simulate calls to effectively search. 
Problems for which actions strongly influence state transition tend to require a larger $\Lambda$.
Normalizing the reward and variance trace to be within the range $[-1,1]$ was found to improve search efficiency. 
Problems in which information gathering is important tend to benefit from including higher $\lambda$ values.

\section{Experiments} \label{sec:Experiments}
We tested the performance of PA-POMCPOW on two tasks involving sensor placement and wildfire containment.
The sensor placement task has a static environment state and a dense reward.
The wildfire containment task has a dynamic environment state and a sparse reward. 
All experiments were implemented using POMDPs.jl~\cite{egorov2017}.
\subsection{Sensor Placement}
The sensor placement task requires an agent to sequentially pick the location to install sensors in a large 2D grid world in order to maximize information gathered. 
Each location in the world has a different concentration of information. 
The information densities are generated by sampling from a Gaussian Process prior with zero mean and a linear-exponential covariance kernel~\cite{kochenderfer2019} at each coordinate. 

The state space is $\mathcal{S} = (\mathcal{S}_g, \mathcal{S}_s)$, where $\mathcal{S}_g$ is the information grid map and $\mathcal{S}_s$ is a list of the coordinates of the placed sensors.
The information field is static throughout each episode, and sensor placement is deterministic.
The field is initialized with a set of sensors placed at random points on the grid, and the values at these points are known.

At each step, the agent chooses the location to place a sensor or to observe. 
The agent can choose a point on the grid that are at least $\delta$ cells away from a previously placed sensor.
If a sensor is placed, the agent receives a reward equal to the value of information at that cell minus one. 
No reward is given for observing without placing a sensor.
Because the agent may either observe or place a sensor at each grid cell, the size of the action space is equal to twice the number of grid cells minus the number of prohibited spaces. 
We considered grids of size $(25\times25)$, $(50\times50)$, $(100\times100)$, corresponding to maximum action space sizes of $1250$, $5000$, and $20000$, respectively.

Each time the agent takes an action, it directly observes the value of the cell on which the sensor is placed. 
The episode terminates after $T$ sensors have been placed.

To solve this task, we used a Gaussian Process to model the belief. 
At each step, the belief is updated by appending the observed location and values to the GP parameters.

We tested PA-POMCPOW by solving the problem for the three different grid sizes with 100 different initializations for each. 
The $\Lambda$ vector was set to linearly spaced values between 0 and 2 with a step size of 0.1 for a total of 20 considered actions. 
Because our belief was Gaussian, we used the exact linear-Gaussian forms of the action-score function.

We ran each test with limits of \num{100}, \num{500}, and \num{1000} 
simulator calls per step.
For each run, we recorded the total accumulated reward and the average planner run time. 
For comparison, we ran each test for the POMCP and POMCPOW algorithms as well, using the same belief distribution and number of solver calls.
We also defined a simple greedy policy in which the agent takes the action with the maximum expected score as defined by the Gaussian Process posterior. 
This was included to test whether limiting the actions of PA-POMCPOW limited the learned policy to a myopic one. 

To test our hypothesis that PA-POMCPOW builds deeper search trees, we measured the maximum tree depth produced by each algorithm. 
For each solver, we constructed the initial action step tree using 500 simulations over 100 state and initial belief realizations. 

\subsection{Wildfire Containment}

In the wildfire containment task, a fire is spreading over a grid world.
An agent must select areas in the grid to clear of fuel in order to contain the spread of the wildfire. 
Fire propagation is probabilistic, according to dynamics defined in a previous work~\cite{julian2019}. 
This task was designed to test the performance of PA-POMCPOW on a task with non-stationary dynamics and a sparse reward.

In this model, fire starts at some points in the grid and burns at those points until the fuel is exhausted. 
A fuel-containing cell that is currently not burning ignites with probability proportional to the number of its neighbors currently burning.
The fire is able to spread beyond immediately adjacent cells, up to two cells away.
Wind biases the fire propagation direction and changes randomly each step.
\begin{figure}[h]
    \centering
    \includegraphics[width=0.49\columnwidth]{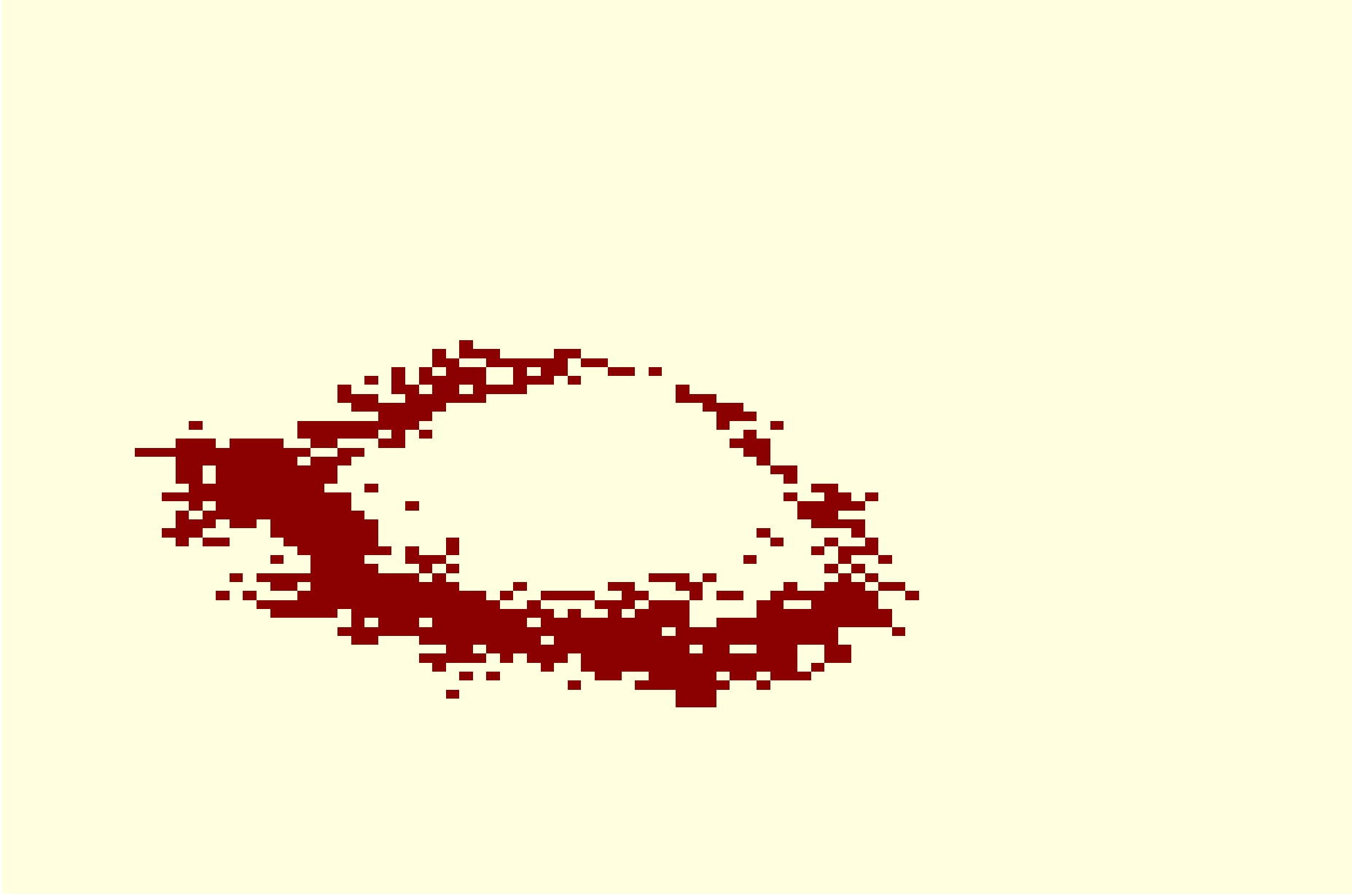}
    \includegraphics[width=0.49\columnwidth]{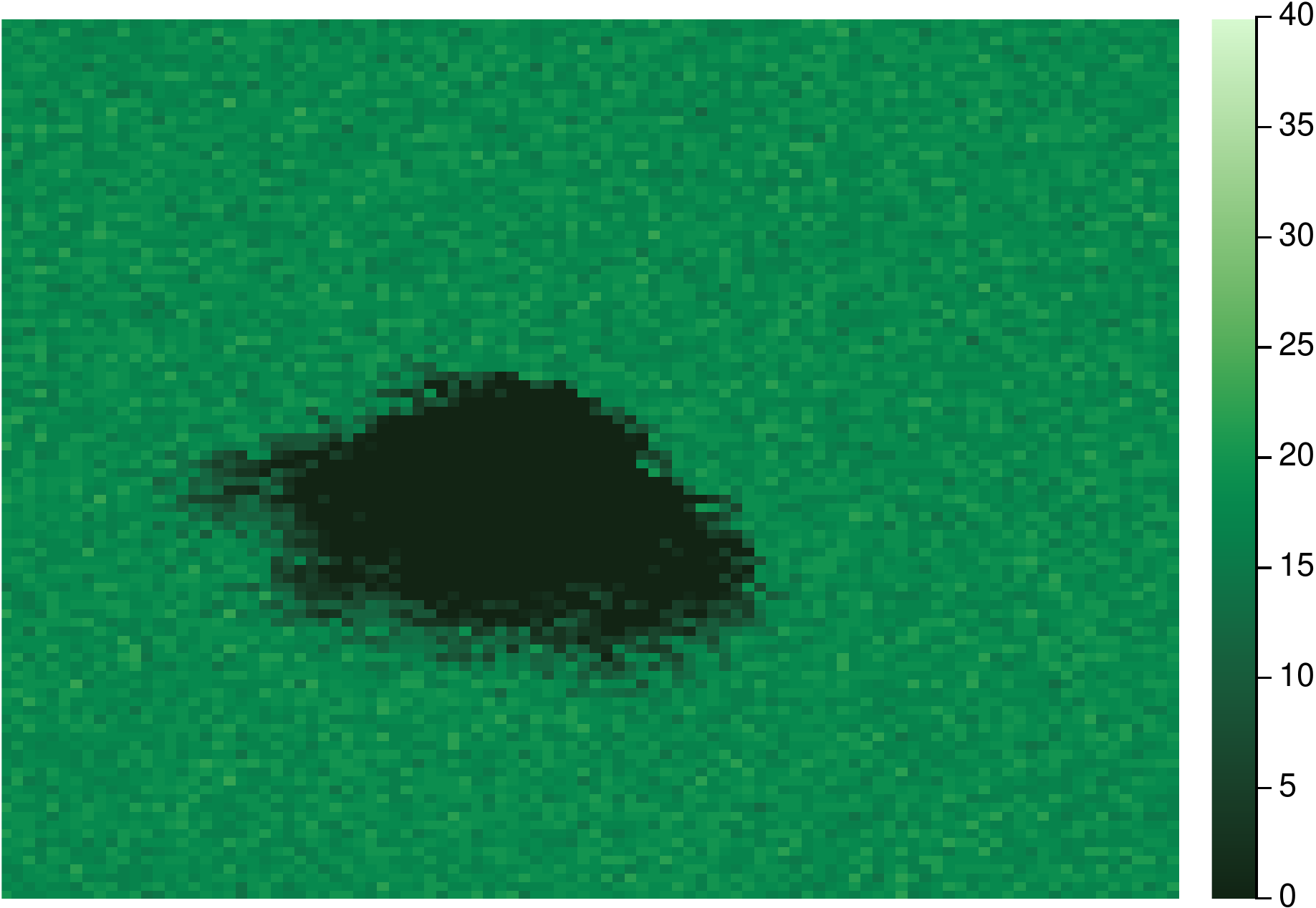}
    \caption{Wildfire containment task. The left figure shows the burn map, where dark areas are currently burning and light are un-ignited. The right figure shows the fuel map, where lighter values correspond to higher fuel levels. A southwest wind biases fire propagation.}
    \label{fig:wildfire}
\end{figure}

The state is represented by a burn map, a fuel map, and a wind vector, as shown in~\Cref{fig:wildfire}.
The burn map is an array of which cells in the grid world are on fire. 
The fuel map is an array of how much fuel is contained in each cell and is generated by sampling each cell from a truncated Gaussian distribution.
Wind is uniform over the grid and the vector is sampled from a 2D uniform distribution between $[-1,1]$.

Areas in each corner of the grid are designated to be keep-out areas. 
Associated with each area is a counter $c_i$ which decreases at each time step until it reaches zero.
The objective of the task is to keep the fire from reaching any keep out zone until the zone counter is zero.
If fire reaches a keep-out zone, that zone's counter is set directly to zero and a reward equal to the remaining counter value, $-c_i$.
The episode terminates when all zones have a zero count. 

At each step, the agent selects a non-burning grid cell to clear of fuel. 
The fuel level in the selected cell and eight surrounding cells are then set to zero. 
The wind vector is updated by addition of zero-mean Gaussian noise.

The agent has full knowledge of the burn map, fuel map, and keep-out counters.
The agent also makes noisy measurements of the wind. 
The noise on the wind measurement is proportional to the distance between the action location and the fire.
Fire reaching any keep-out zone has a cost equal to ten times the counter value of the keep-out zone. 

To solve this task, we used a Gaussian distribution to model the wind belief,  and updated it using a Kalman filter~\cite{kalman1960}.
The $\Lambda$ set was composed of linearly spaced values between 0.5 and 1.5 with a step size of 0.1 for a total of 16 values. 
The linear-Gaussian forms of the score function were used, however, because the reward was sparse, a dense, shaped reward was implemented.
The shaped reward function was defined as $r(a,s) = \theta d_f(a) + \beta d_k(a)$, where $d_f$ measures the distance of the cleared cell to the closest burning cell and $d_k(a)$ is the distance to the nearest keep-out zone.
The $\theta$ and $\beta$ terms are weighting factors.

We ran each test with limits of $100$, $250$, and $500$ simulator calls per solver step for a grid size of $40\times40$.
For each run, we recorded the total accumulated reward and the average planner run time per step. 
As with the sensor placement task, we ran each test for the POMCP and POMCPOW baseline algorithms as well, using the same belief distribution and number of solver calls.

We additionally implemented a myopic expert policy baseline that clears all the fuel bordering each keep-out zone. 
At each step, the policy clears the cells immediately bordering a keep-out zones in order to create a barrier around the zone.
The policy chooses to clear cells of the keep-out zone that is closest to the fire, until all the bordering cells have been cleared. 
It then moves to the next closest zone. 

\section{Results} \label{sec:Results}
\subsection{Sensor Placement}
\begin{figure*}[t!]
    \centering
    \begin{subfigure}[t]{1.6in}
		\centering
		\includegraphics[width=1.55in]{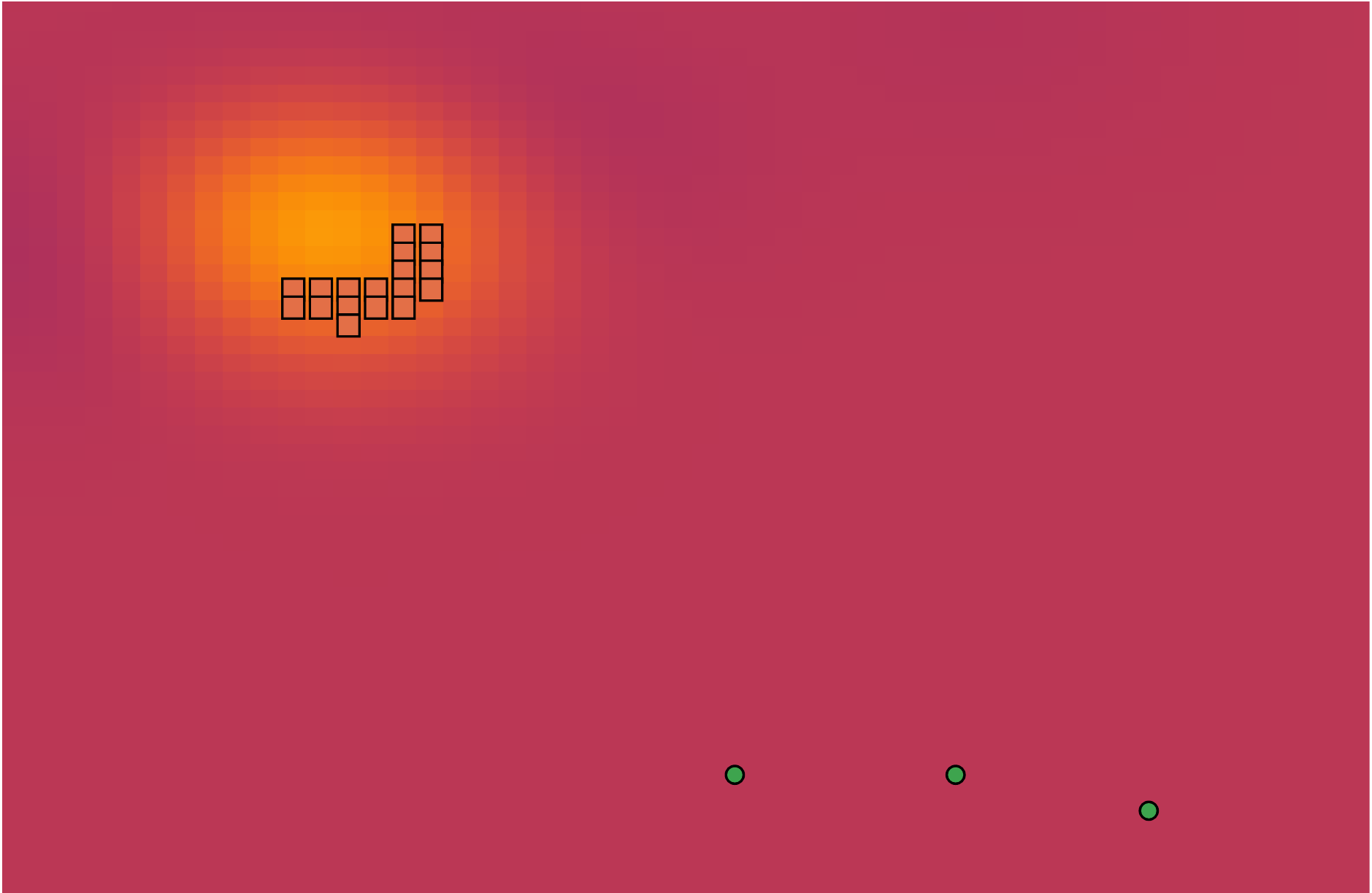}
		\caption{}\label{fig:means(a)}
	\end{subfigure}
    \begin{subfigure}[t]{1.6in}
		\centering
		\includegraphics[width=1.55in]{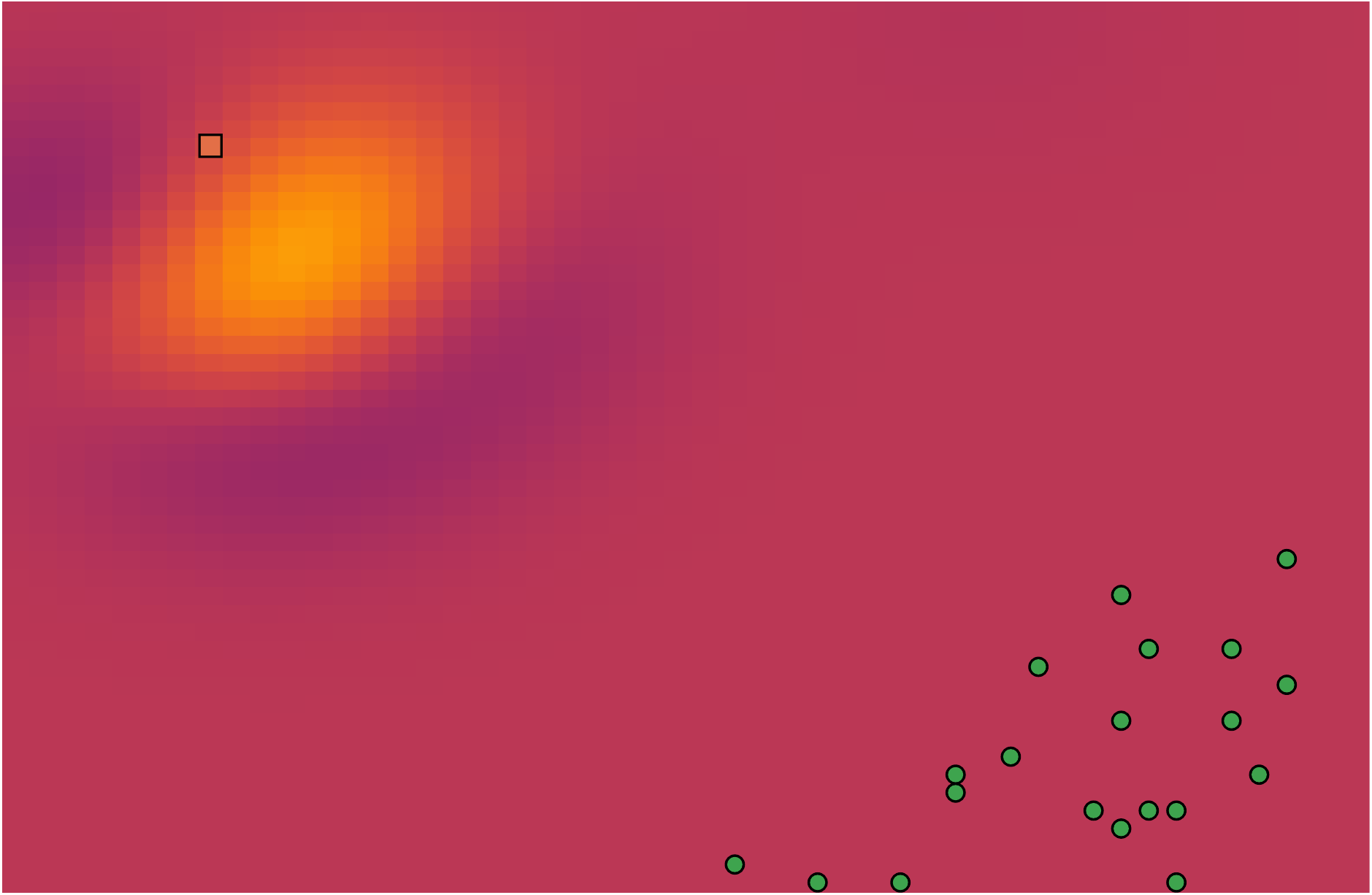}
		\caption{}\label{fig:means(b)}
	\end{subfigure}
	\begin{subfigure}[t]{1.6in}
		\centering
		\includegraphics[width=1.55in]{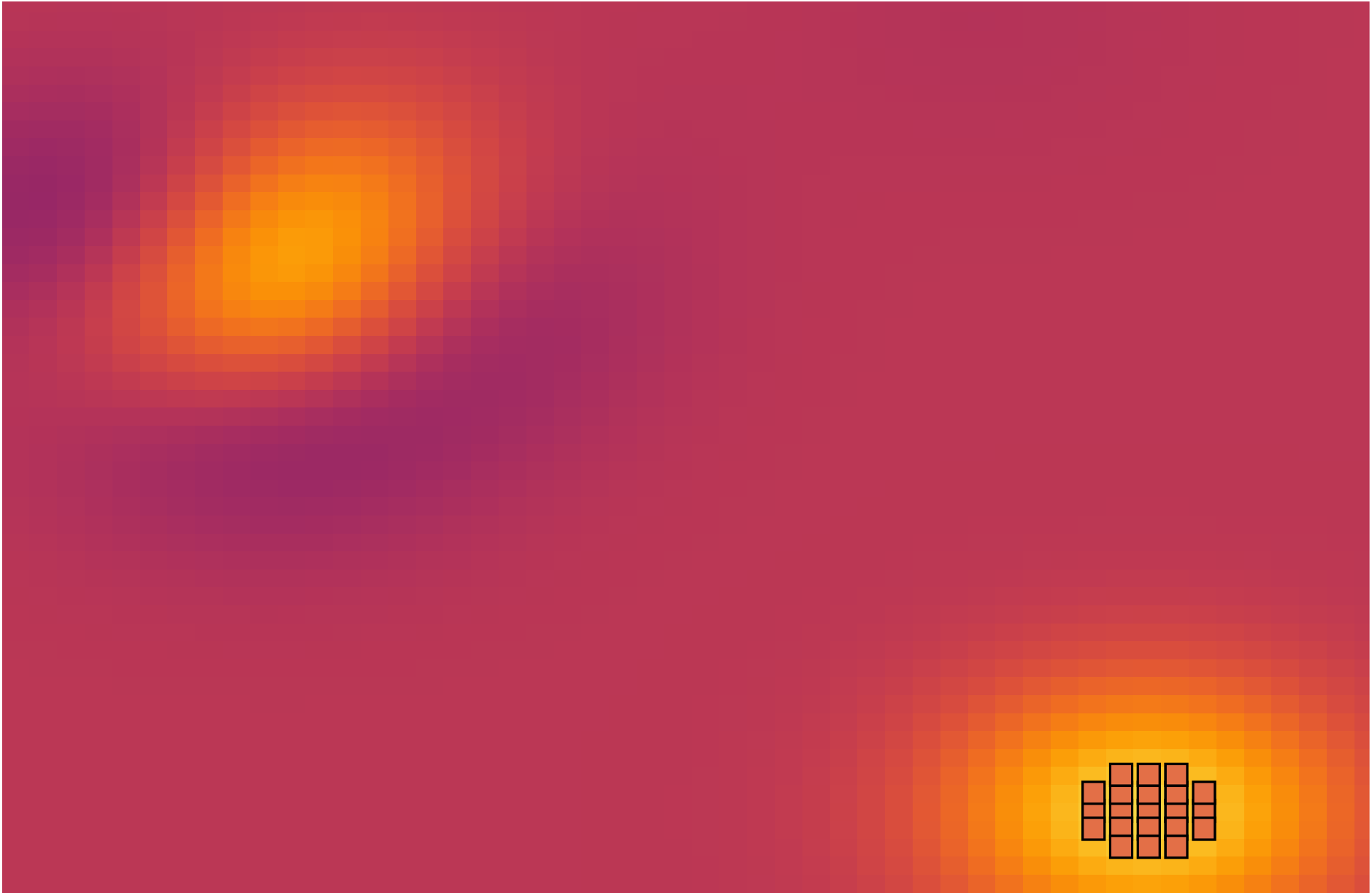}
		\caption{}\label{fig:means(c)}
	\end{subfigure}
	\begin{subfigure}[t]{1.6in}
		\centering
		\includegraphics[width=1.6in]{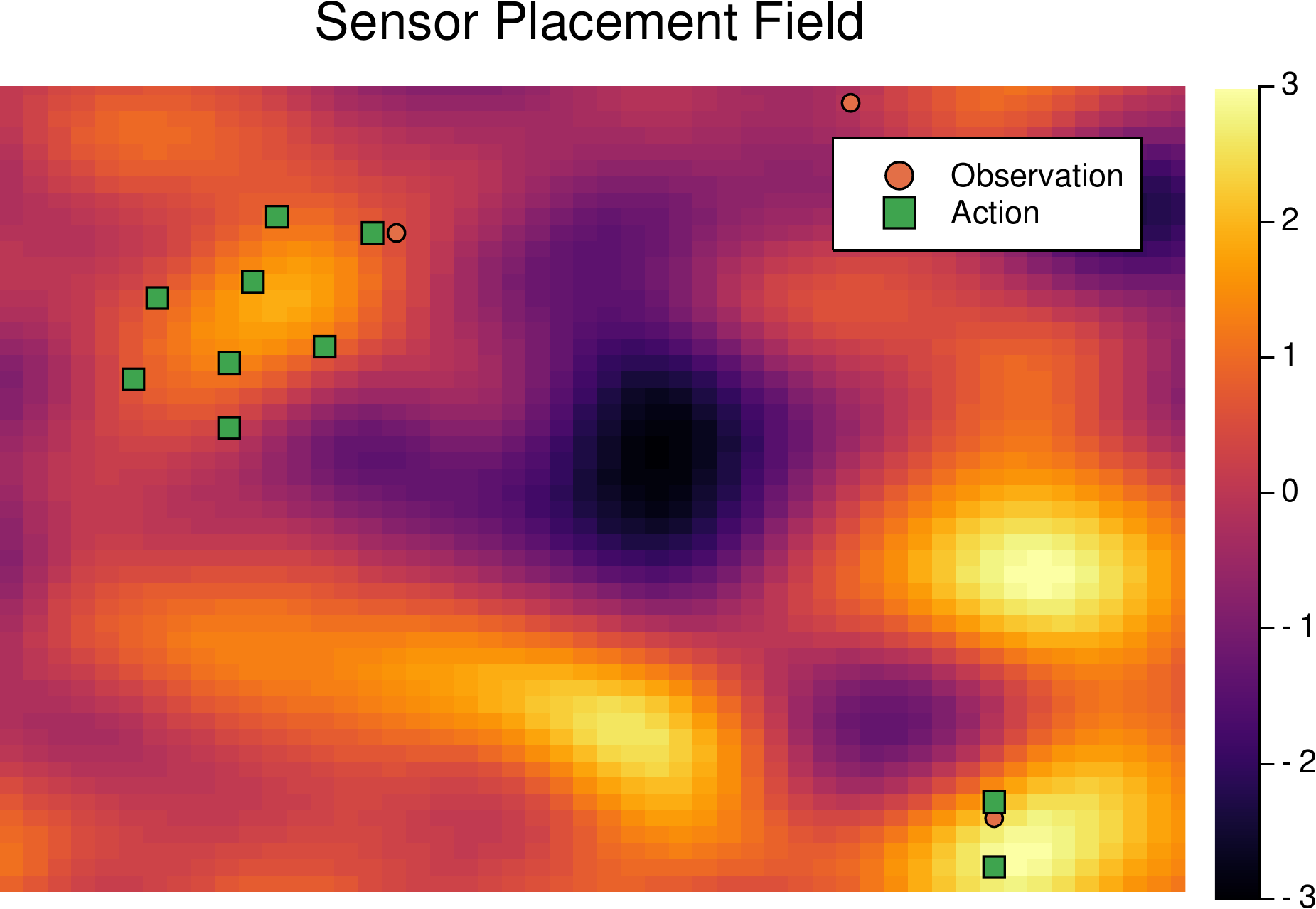}
		\caption{}\label{fig:means(d)}
	\end{subfigure}
    \caption{Example sensor placement action branching. The first three figures show the Gaussian process mean values with root node actions overlaid. The squares mark sensor placement actions and the circles mark observation only actions. (a) The algorithm prefers actions with high known reward early in the episode. (b) As the rewarding actions are depleted, the algorithm considers more exploration. (c) The algorithm once again prefers high-reward actions once a new high-value area is found. (d) The actual information state for the episode and selected actions at episode completion.}
    \label{fig:sensor_exmp}
\end{figure*}{}
The performance of each algorithm on the sensor placement task is reported in~\Cref{table:sensor_score}.
The mean score and standard error over the 100 trials are reported for each test point. 

\begin{table*}[!tb]
    \begin{minipage}{.65\linewidth}
    \caption{Sensor Placement Task Scores}\label{table:sensor_score}
      \centering
\resizebox{\textwidth}{!}{
 \begin{tabular}{@{}llcccc@{}} 
    \toprule
    Grid Size & Calls & \textbf{PA-POMCPOW} & POMCPOW & POMCP & Greedy \\
    \midrule
    \multirow{3}{*}{$20\times20$} & $100$ &  $3.35 \pm 0.22$ & $ 0.77\pm 0.22$ & $1.31 \pm 0.24$ & \multirow{3}{*}{$2.55 \pm 0.21$} \\
    & $500$ & $3.66 \pm 0.21$ & $1.45 \pm 0.23$ & $1.26 \pm 0.20$ &\\
    & $1000$ & $\mathbf{4.04 \pm 0.21}$ & $1.70 \pm 0.26$ & $1.18 \pm 0.23$ &\\
    \midrule
    \multirow{3}{*}{$50\times50$} & $100$ &  $\mathbf{5.79 \pm 0.29}$ & $0.42 \pm 0.24$ & $1.40 \pm 0.25$ & \multirow{3}{*}{$5.10 \pm 0.32$}\\
    & $500$ & $5.64 \pm 0.28$ & $2.90 \pm 0.28$ & $0.96 \pm 0.23$ & \\
    & $1000$ & $5.46 \pm 0.29$ & $3.80 \pm 0.38$ & $0.48 \pm 0.26$ & \\
    \midrule
    \multirow{3}{*}{$100\times100$} & $100$ &  $6.45 \pm 0.44$ & $3.99 \pm 0.41$ & $3.36 \pm 0.44$ & \multirow{3}{*}{$6.69 \pm 0.40$} \\
    & $500$ & $7.68 \pm 0.41$ & $5.64 \pm 0.43$ & $3.10 \pm 0.41$ & \\
    & $1000$ & $\mathbf{7.77 \pm 0.44}$ & $5.57 \pm 0.43$ & $3.10 \pm 0.39$ & \\
    \bottomrule
\end{tabular}}
    \end{minipage}%
    \begin{minipage}{.35\linewidth}
      \centering
        \caption{Wildfire Task Loss}\label{table:wildfire}
        \resizebox{0.8\textwidth}{!}{
        \begin{tabular}{@{}llc@{}} 
        \toprule
        Calls & Algorithm & Loss\\
        \midrule
        \multirow{3}{*}{$100$} & PA-POMCPOW & $\mathbf{460 \pm 46}$ \\
        & POMCPOW & $937 \pm 24$ \\
        & POMCP & $1021 \pm 30$ \\
        \midrule
        \multirow{3}{*}{$250$} & PA-POMCPOW & $\mathbf{434 \pm 46}$ \\
        & POMCPOW & $897 \pm 33$ \\
        & POMCP & $1000 \pm 28$ \\
        \midrule
        \multirow{3}{*}{$500$} & PA-POMCPOW & $\mathbf{430 \pm 43}$ \\
        & POMCPOW & $798 \pm 29$ \\
        & POMCP & $1011 \pm 28$ \\
        \midrule
        -- & Expert & $722 \pm 18$ \\
        \bottomrule
        \end{tabular}}
    \end{minipage} 
\end{table*}

PA-POMCPOW outperformed the baseline algorithms for all test points. 
It also outperformed the Greedy policy in all but one test point, showing that with action subsets, the tree search is still able to find non-myopic policies. 
Neither baseline outperformed the greedy policy in any case. 

The relative gap between POMCPOW and Greedy remained at approximately $30\%$ for $1000$ queries at all grid sizes. 
This seems to suggest that the shallow trees generated by POMCPOW resulted in selection of the approximately greedy action. 
This also suggests why PA-POMCPOW, with deeper search trees, was able to outperform it. 

On average, PA-POMCPOW required \SI{4.2}{\milli\second}, \SI{8.3}{\milli\second}, and \SI{43.5}{\milli\second} per simulation call for the $20\times20$ grid, $50\times50$ grid, and $100\times100$ grid respectively.
This was significantly more expensive than the most efficient baseline, POMCPOW, which required \SI{0.7}{\milli\second}, \SI{2.4}{\milli\second}, \SI{10.1}{\milli\second} for the three respective grids.
This is likely due to the added cost of computing the action scores at each observation node.

A partial episode is shown in~\Cref{fig:sensor_exmp}.
The location and type of each root node action of the search tree is shown overlaid on the Gaussian process belief mean values. 
A balance of exploration and exploitation is seen over the episode, however, when high expected reward actions are available, the actions tend to form tight clusters, which may not be desirable in some tasks. 

From the tree measurement experiment, the average maximum node depth and standard error was $8.13 \pm 0.23$, $3.61 \pm 0.06$, and $3.05 \pm 0.05$ for PA-POMCPOW, POMCPOW, and POMCP, respectively. 
The average maximum depth was significantly higher for PA-POMCPOW than either baseline. 
These results suggest that improved tree depth contributed to PA-POMCPOW's improved performance on the task. 

\subsection{Wildfire Containment}
The performance results from the wildfire containment task are shown in~\Cref{table:wildfire}.
As with the sensor placement task, PA-POMCPOW was able to outperform both of the baseline algorithms and the expert policy in all three test scenarios. 
The expert policy outperformed both baselines. 

The computational cost of the wildfire task was slightly higher than that of the sensor placement task. 
The time per-query was \SI{11.6}{\milli\second} for PA-POMCPOW, \SI{8.5}{\milli\second} for POMCPOW, and \SI{8.2}{\milli\second} for POMCP. 
As before, PA-POMCPOW was more expensive than POMCPOW and POMCP.

Despite the more complex environment and sparse reward function, PA-POMCPOW was still able to solve the problem better than the existing state-of-the-art and an expert policy. 

\section{Conclusions} \label{sec:Conclusions}
We presented a general method to extend online POMDP solvers to problems with very large action spaces by prioritizing actions for tree expansion.
Specific formulations of the method for various reward functions and belief distributions were presented.
We implemented this method as a new algorithm called Prioritized Action POMCPOW (PA-POMCPOW) which can scale to very large problems. 

The current work is limited to problems in which the score function terms can be analytically formed or easily and accurately approximated.
Future work will investigate more generally extensible functions for the exploration and exploitation terms of the action score. 

This work presented a static method of selecting the action subset. 
That is, once selected, the action space subset was never updated. 
Future work will explore dynamically adjusting the subset based on the action-value estimates. 

This work also only directly considered large, discrete action spaces.
While the proposed methods can be applied to continuous spaces in principle, evaluating the action score in a continuous domain would likely be intractable for many problems. 
Because of this, future work will investigate using the volume coverage metrics for action clustering~\cite{kurniawati2008}.

Despite these limitations, experimental results showed that PA-POMCPOW was effective on very large problems. 
Using the proposed method with DPW improved the performance over existing state-of-the-art methods. 

\bibstyle{aaai21} 
\bibliography{bibliography}
\end{document}